\documentclass{article}
\usepackage{spconf,amsmath,graphicx}
\usepackage{xcolor}
\usepackage{threeparttable}
\usepackage{multirow}
\usepackage{booktabs}
\usepackage{caption}
\usepackage{subcaption}
\usepackage{enumitem}
\setlist{nosep, leftmargin=14pt}
\usepackage{url}

\title{Inspecting Model Fairness in Ultrasound Segmentation Tasks}
\name{Zikang Xu$^{1,2}$, Fenghe Tang$^{1,2}$, Quan Quan$^{3}$, Jianrui Ding$^{4}$, Chunping Ning$^{5}$, S. Kevin Zhou $^{1,2,3,\star}$}
\address{
    $^{1}$ School of Biomedical Engineering, Division of Life Sciences and Medicine, \\ University of Science And Technology of China, Suzhou, China\\
    $^{2}$ Suzhou Institute for Advanced Research,  University of Science And Technology of China, Suzhou, China\\
    $^{3}$ Key Lab of Intelligent Information Processing of Chinese Academy of Sciences (CAS),\\ Institute of Computing Technology, CAS, Beijing, China\\
    $^{4}$ School of Computer Science and Technology, Harbin Institute of Technology, Harbin, China\\
    $^{5}$ Ultrasound Department, The Affiliated Hospital of Qingdao University, Qingdao, China\\
}

\begin{document}
%
\maketitle
\begin{abstract}

With the rapid expansion of machine learning and deep learning (DL), researchers are increasingly employing learning-based algorithms to alleviate diagnostic challenges across diverse medical tasks and applications. While advancements in diagnostic precision are notable, some researchers have identified a concerning trend: their models exhibit biased performance across subgroups characterized by different sensitive attributes. This bias not only infringes upon the rights of patients but also has the potential to lead to life-altering consequences.
In this paper, we inspect a series of DL segmentation models using two ultrasound datasets, aiming to assess the presence of model unfairness in these specific tasks. Our findings reveal that even state-of-the-art DL algorithms demonstrate unfair behavior in ultrasound segmentation tasks. These results serve as a crucial warning, underscoring the necessity for careful model evaluation before their deployment in real-world scenarios. Such assessments are imperative to ensure ethical considerations and mitigate the risk of adverse impacts on patient outcomes.
\end{abstract}

\section{Introduction}
\label{sec:intro}


In recent years, the widespread adoption of machine learning and deep learning (DL) has played a pivotal role in assisting medical professionals with a spectrum of tasks related to the analysis of medical images. These tasks range from fundamental processes such as reconstruction and denoising to more intricate endeavors including classification, segmentation, and detection~\cite{zhou2021review}.

The integration of these advanced methodologies enables doctors to derive disease predictions directly from raw medical images, leading to a reduction in workload, improved work efficiency, and a significant enhancement in diagnostic precision. The versatility of DL methods has been successfully demonstrated across various imaging modalities, encompassing X-Ray~\cite{baltruschat2019comparison}, Magnetic Resonance Imaging (MRI)~\cite{akkus2017deep}, Computed Tomography (CT)~\cite{larrazabal2020gender}, Ultrasound (US)~\cite{cmunet}, and others. This broad applicability underscores the transformative impact of DL in diverse medical imaging applications, promising continued advancements in the realms of healthcare diagnostics and treatment planning.

However, there are also some unobtrusive shortcomings of DL models in medical applications. One among them is model \textbf{unfairness}~\cite{glocker2022risk}, which is defined as the utility disparity of DL models when applied to sub-populations with different sensitive attributes, such as \textit{Age}, \textit{Sex}, \textit{Race}, and \textit{Marital}. This phenomenon occurs widely in medical classification tasks.

Nonetheless, there are inconspicuous limitations in the application of DL models to medical scenarios. One notable concern is the issue of unfairness in models, characterized by a discernible discrepancy in their utility when employed across sub-populations distinguished by different sensitive attributes, such as age, sex, race, or marital status. This phenomenon is notably prevalent in medical classification tasks.

For instance, Larrazaba~\textit{et al.}~\cite{larrazabal2020gender} observed substantial performance variations among three DL models when applied to distinct gender groups (male and female) in a chest X-ray diagnosis task. Similarly, researchers have identified this phenomenon in skin lesion detection tasks, revealing that state-of-the-art DL models exhibit uneven performances between patients with dark and light skin tones~\cite{deng2023fairness,xu2023fairadabn}.


In recent times, researchers have turned their attention to the critical task of evaluating fairness in segmentation processes. Much of this research has concentrated on brain MR images, uncovering instances of model unfairness, particularly in brain tumor segmentation~\cite{lee2023investigation}. The underlying reason for this phenomenon lies in the relative ease with which DL models can predict the sex and age of subjects in MR images compared to directly predicting Alzheimer's state. Consequently, models may inadvertently exploit spurious correlations between sex and the target label, resulting in observed unfairness~\cite{zong2022medfair}.

Comparatively, obtaining sex and age information from US images is inherently more challenging than from MR images. Furthermore, as of our current knowledge, there is a lack of evidence confirming the absence of unfairness in US segmentation tasks. Therefore, it is imperative to scrutinize the potential existence of unfairness issues in US segmentation. This examination is crucial not only for validating the origin of unfairness but also for alerting patients to the necessity of careful consideration before embracing DL models for US segmentation applications.

Our contributions to the field are outlined as follows:

(1) We conduct the \textbf{first} comprehensive assessment of model fairness in US segmentation tasks. This evaluation encompasses seven state-of-the-art U-Net based segmentation algorithms.

(2) Through rigorous statistical analysis, we identify varying degrees of unfairness within the state-of-the-art segmentation methods. This underscores the importance of acknowledging and addressing model unfairness as a critical consideration before the deployment of these methods.

\begin{table}[t]
    \centering
    \begin{tabular}{cccc}
        \toprule
          \multicolumn{2}{c}{Dataset} &TUSC &  TUS \\
      \midrule
          \multirow{2}{*}{Age} & Young & 479 (55.70\%) & 884 (56.23\%) \\
                               & Old & 381 (44.30\%) & 688 (43.77\%) \\
      \midrule
          \multirow{2}{*}{Sex} & Female & 722 (83.95\%) & 1331 (84.67\%) \\
                               & Male & 138 (16.05\%) & 241 (15.33\%)\\    
      \bottomrule
    \end{tabular}
    \caption{Dataset distribution of TUSC and TUS}
    \label{tab:dataset_distribution}
\end{table}

\begin{table*}[h]
    \centering
    \resizebox{\linewidth}{!}{
    \begin{tabular}{rccccccc|ccccccc}    
        \toprule
        \multicolumn{8}{c}{\textbf{TUSC Dataset}} & \multicolumn{7}{|c}{\textbf{TUS Dataset}} \\
        \midrule
        \multirow{2}{*}{Method} & \multirow{2}{*}{Dice $\uparrow$}  & \multicolumn{3}{c}{Sex (\%)} & \multicolumn{3}{c|}{Age (\%)} & \multirow{2}{*}{Dice $\uparrow$}  & \multicolumn{3}{c}{Sex (\%)} & \multicolumn{3}{c}{Age (\%)} \\
        \cmidrule(r){3-5} \cmidrule(r){6-8} \cmidrule(r){10-12} \cmidrule(r){13-15}
          & &  ESSP $\uparrow$& DD $\downarrow$ & EOpp $\downarrow$ & ESSP $\uparrow$ & DD $\downarrow$& EOpp $\downarrow$ & & ESSP $\uparrow$& DD $\downarrow$ & EOpp $\downarrow$ & ESSP $\uparrow$ & DD $\downarrow$& EOpp $\downarrow$ \\
        \midrule
        
        \textit{{\color{gray} CNN-based}} &&&&&&&&& \\
        U-Net~\cite{unet}	&	67.57	$\pm$	0.71	&	67.47	&	\underline{0.30}	&	4.06	&	66.20	&	4.15	&	\underline{3.32}	& 82.06	$\pm$	0.44	&	81.88	&	0.44	&	0.44	&	81.63	&	1.05	&	2.18	\\
        UNeXt~\cite{unext}	&	63.56	$\pm$	1.31	&	63.51	&	\textbf{0.17}	&	\textbf{0.83}	&	62.56	&	3.19	&	\textbf{1.93}	&	79.14	$\pm$	0.53	&	78.49	&	1.65	&	\textbf{0.13}	&	78.71	&	1.11	&	1.74	\\
        U-Net++~\cite{unet++}	&	\underline{71.78	$\pm$	0.36}	&	\underline{71.19}	&	1.67	&	1.56	&	\underline{70.90}	&	\underline{2.50}	&	4.40&	81.72	$\pm$	0.68	&	81.68	&	\underline{0.10}	&	1.25	&	81.45	&	\underline{0.67}	&	1.83	\\
        Attention U-Net~\cite{attunet}	&	68.22	$\pm$	0.87	&	67.74	&	1.40	&	3.24	&	66.80	&	4.24	&	3.54&	\underline{82.19	$\pm$	0.42}	&	\underline{82.09}	&	0.25	&	0.98	&	\underline{81.75}	&	1.07	&	\textbf{1.27}	\\
        CMU-Net~\cite{cmunet}	&	\textbf{74.41	$\pm$	1.26}	&	\textbf{74.09}	&	0.84	&	\underline{1.04}	&	\textbf{73.57}	&	\textbf{2.26}	&	3.82&	\textbf{82.54	$\pm$	0.46}	&	\textbf{82.35}	&	0.47	&	\underline{0.27}	&	\textbf{81.95}	&	1.43	&	2.31	\\
        CMUNeXt~\cite{cmunext}	&	68.07	$\pm$	0.74	&	67.37	&	2.07	&	1.28	&	66.65	&	4.26	&	4.12&	81.94	$\pm$	0.34	&	81.90	&	\textbf{0.10}	&	1.02	&	81.61	&	0.81	&	1.68	\\
        \textit{{\color{gray} Transformer-based}} &&&&&&&&& \\
        TransUnet~\cite{transunet}	&	68.75	$\pm$	1.29	&	68.38	&	1.07	&	1.16	&	67.13	&	4.80	&	5.87&	81.66	$\pm$	0.18	&	81.59	&	0.15	&	-0.12	&	81.44	&	\textbf{0.52}	&	\underline{1.59}	\\
         \bottomrule
    \end{tabular}
    }
    \caption{Performance and fairness scores on \textit{Sex} and \textit{Age} attribute. The \textbf{best} and \underline{second best} are highligted.}
    \label{tab:sex}
\end{table*}

\section{Method}
\label{sec:method}

\subsection{Problem Definition}

Considering a medical segmentation task, where the input dataset $D$ consists of $N$ samples, \textit{i.e.} $D=\{s_1, s_2, ..., s_N\}$. Each sample $s_i = \{X_i, Y_i, A_i\}$ is a tuple of the input image $X_i\in R^{C \times W \times H}$, the ground truth mask $Y_i\in R^{C \times W \times H}$, and a \textbf{sensitive attribute} $A_i\in\{0, 1\}$, which usually is a binary variable that describes the metadata of the sample, such as \textit{Age}, \textit{Sex}, and \textit{Race}.
In group fairness settings, the performance of the DL model is computed per sample and averaged on the groups split by sensitive attributes. 
Ideal group fairness requires that different subgroups to have the same performance, which is hard to achieve. Thus, the performance disparity is computed across subgroups to measure the unfairness.

\subsection{Experiment Settings}



In this paper, we conduct a group fairness evaluation on two US datasets: the publicly available Thyroid UltraSound Cine-clip dataset~\cite{TUSC} (TUSC) and a private Thyroid Ultrasound dataset (TUS). Both datasets include sex and age information for the patients, and their distributions are presented in the table.

For the variable \textit{Age}, treated as a continuous variable, we adopt a threshold of 60 for the Age subgroup split in TUSC, while for TUS, we simplify the threshold to 45 years old, following the processing outlined in~\cite{deng2023fairness}.
The TUSC dataset comprises ultrasound cine-clip images from 167 patients gathered at Stanford University Medical Center. Due to slight differences in the images extracted from the ultrasound video, we resample the dataset at a ratio of 5, resulting in a dataset of 860 images.
The TUS dataset is collected from the Ultrasound Department of the Affiliated Hospital of Qingdao University, consisting of 1,942 images from 192 cases, annotated by three experienced radiologists. For both datasets, we randomly partition them into training and testing sets with a ratio of 7:3.

Our evalutations are done with seven state-of-the-art image segmentation methods, including CNN-baseds network such as: U-Net~\cite{unet}, U-Net++~\cite{unet++}, UNeXt~\cite{unext},  Attention U-Net~\cite{attunet}, CMU-Net~\cite{cmunet}, and CMUNeXt~\cite{cmunext}, and transformer-based networks TransUnet~\cite{transunet}.

For fair comparison, all of the networks are trained with the same loss $\mathcal{L}$, consisting of binary cross-entropy loss and dice loss, given by the following equation:

\begin{equation}
    \mathcal{L} = \mathcal{L}_{Dice}(y, \hat{y}) + \alpha \mathcal{L}_{BCE}(y, \hat{y})
\end{equation}
where $y$ and $\hat{y}$ denote the ground truth mask and the model prediction, respectively. $\alpha$ is the weight and is set as $\alpha=0.5$.

For utility evaluation, we employ the Dice metric directly. Regarding fairness assessment, drawing inspiration from the concept introduced in~\cite{tian2023fairseg}, we utilize the Equity-Scaled Segmentation Performance (ESSP) and Dice Disparity (DD) to evaluate unfairness in US segmentation. The formulations for ESSP and DD are provided by Equations~\ref{essp} and~\ref{dd}.
\begin{equation}
    \text{ESSP} = \frac{\text{Dice}_{avg}}{1 + \text{Dice}_{std}}
    \label{essp}
\end{equation}

\begin{equation}
    \text{DD} = \text{Dice}_{A=1} - \text{Dice}_{A=0}
    \label{dd}
\end{equation}
where $\text{Dice}_{A=0}$ denotes the average Dice of the unprivileged group (lower performance) and $\text{Dice}_{A=1}$ denotes the average Dice of the privileged group (higher performance).

We also use Equal Opportunity (EOpp)~\cite{hardt2016equality}, which is a commonly used metric designed for fairness evaluation on classification tasks, by regarding the segmentation as pixel-level classification. The formula of EOpp is given by Equ.~\ref{equ:eopp}.

\begin{equation}
    \text{EOpp} = \text{TPR}_{A=1} -\text{TPR}_{A=0}
    \label{equ:eopp}
\end{equation}
where $\text{EOpp}_{A=0}$ denotes the True Positive Rate (TPR) of the unprivileged group (lower performance) and $\text{EOpp}_{A=1}$ denotes the TPR of the privileged group (higher performance).

We resize all training cases of four datasets to 256$\times$256 and apply random rotation and flip for simple image augmentations. In addition, we use the Adam optimizer with a weight decay of 1e-4 and a momentum of 0.9 for training. The initial learning rate is set to 0.01, and the poly strategy is used to adjust the learning rate. The batch size is set to 8 and the training epochs are 300. All the experiments are conducted using 8x NVIDIA GeForce RTX3090Ti GPUs.

\begin{figure*}[h]
    \centering
    \begin{subfigure}{0.49\linewidth}
    \centering
    \caption{Result on \textit{Sex} on TUSC Dataset.}    
    \includegraphics[width=\textwidth]{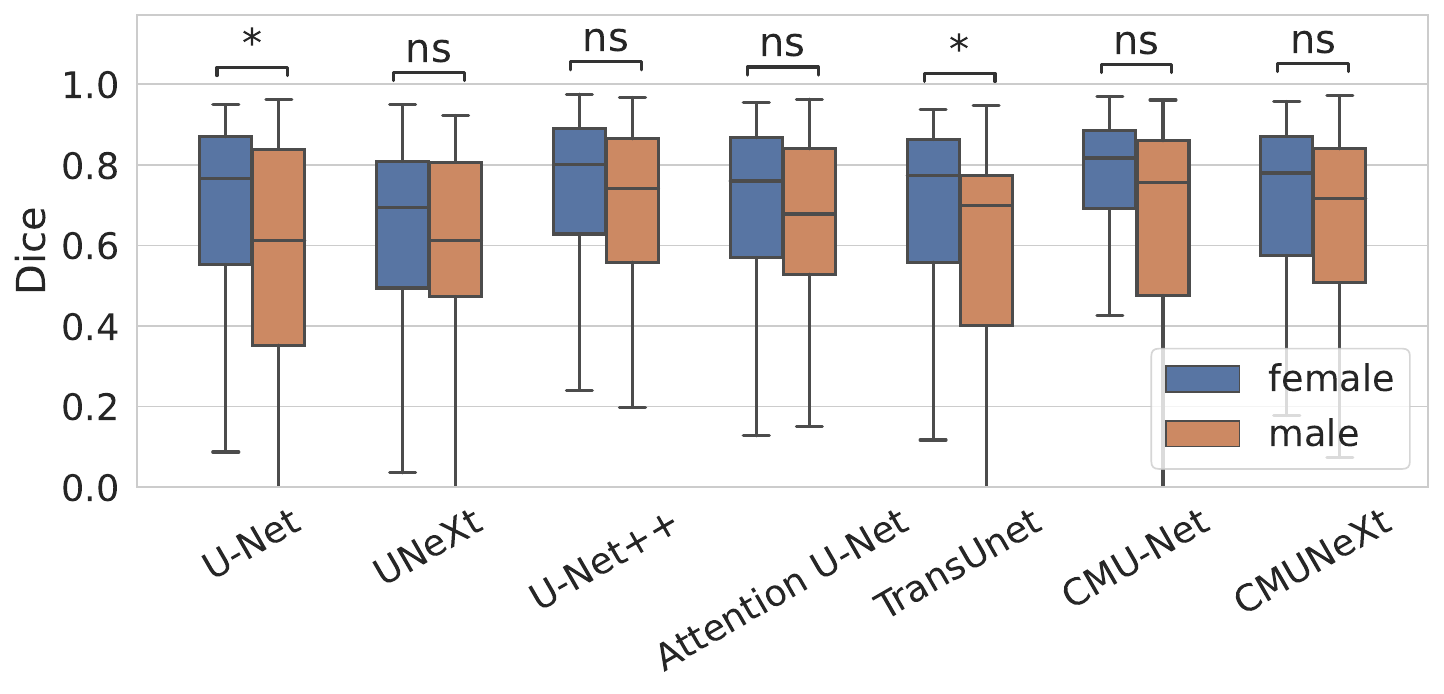}
  \end{subfigure}
    \begin{subfigure}{0.49\linewidth}
    \centering
    \caption{Result on \textit{Age} on TUSC Dataset.}    
    \includegraphics[width=\textwidth]{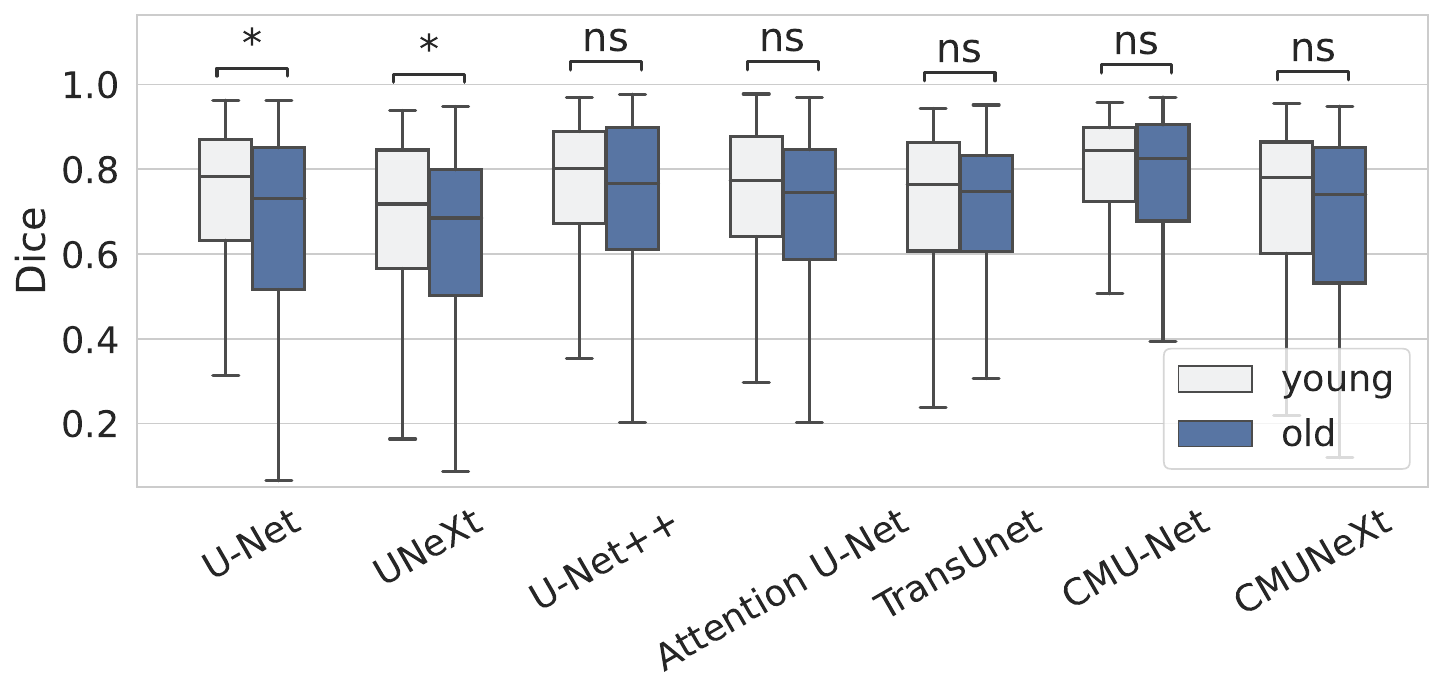}
  \end{subfigure}
  
    \begin{subfigure}{0.49\linewidth}
    \centering
    \caption{Result on \textit{Sex} on TUS Dataset.}    
    \includegraphics[width=\textwidth]{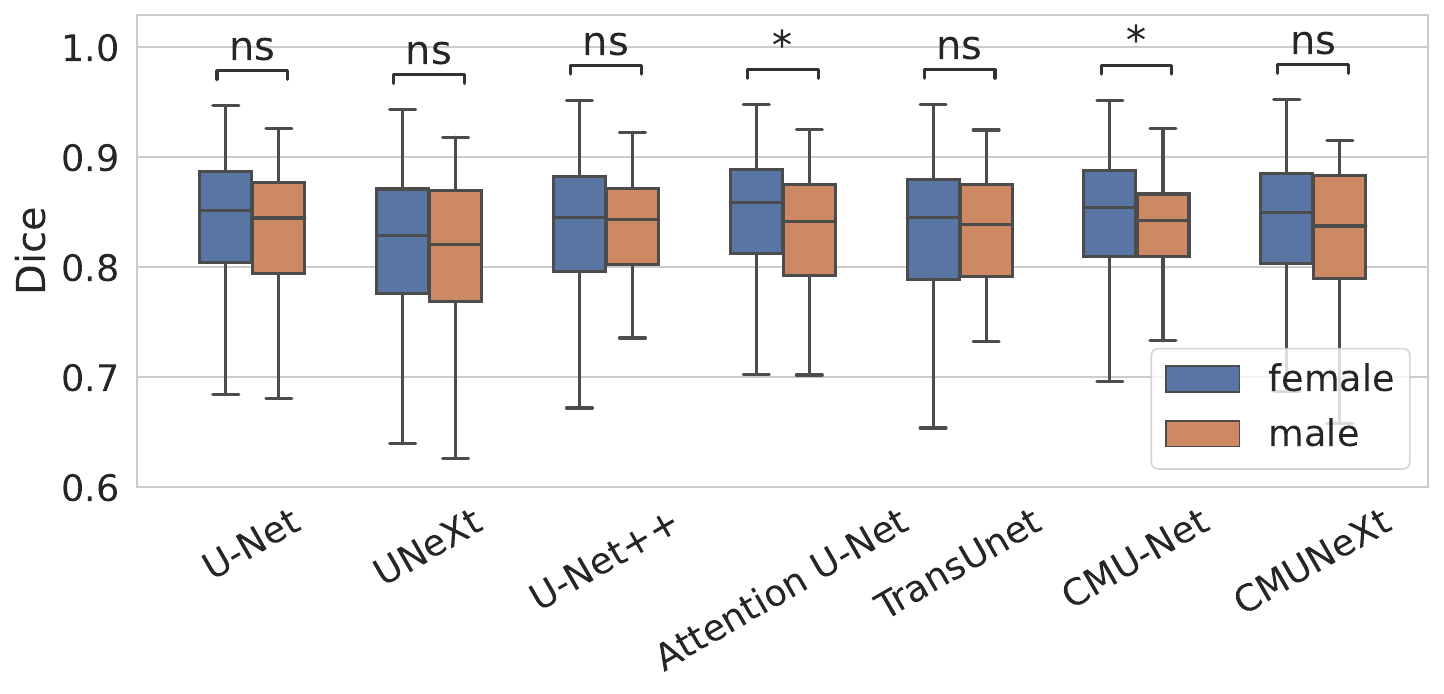}   
  \end{subfigure}
    \begin{subfigure}{0.49\linewidth}
    \centering
    \caption{Result on \textit{Age} on TUS Dataset.}    
    \includegraphics[width=\textwidth]{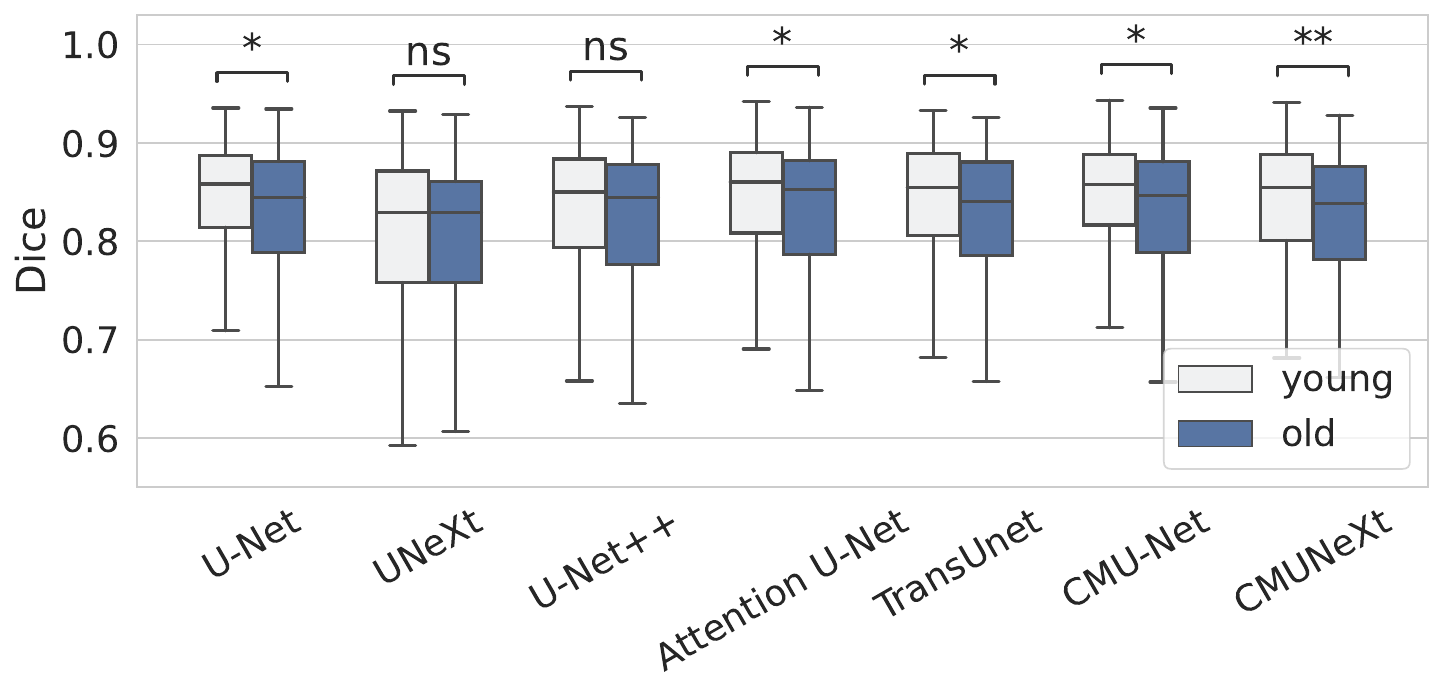}
  \end{subfigure}  
    \caption{Overall Dice for each of the seven models tested. Statistical significance was found using a Mann-Whitney U test and is denoted by **** ( $p\leq$ 0.0001), *** (0.001 $<p\leq$ 0.0001), ** (0.01 $<p\leq$ 0.001), * (0.01 $<p\leq$ 0.05), ns (0.05 $\leq p$ ).}
    \label{fig.1}
\end{figure*}

\section{Result}
\label{sec:result}



In our analysis of fairness regarding predictive performance and model calibration, Table~\ref{tab:sex} presents various metrics illustrating the performance of the DL models for  US diagnosis. The metrics are averaged over the test set across three runs, along with corresponding 95\% confidence intervals.

\begin{figure}[h]
    \centering
    \includegraphics[width=0.7\linewidth]{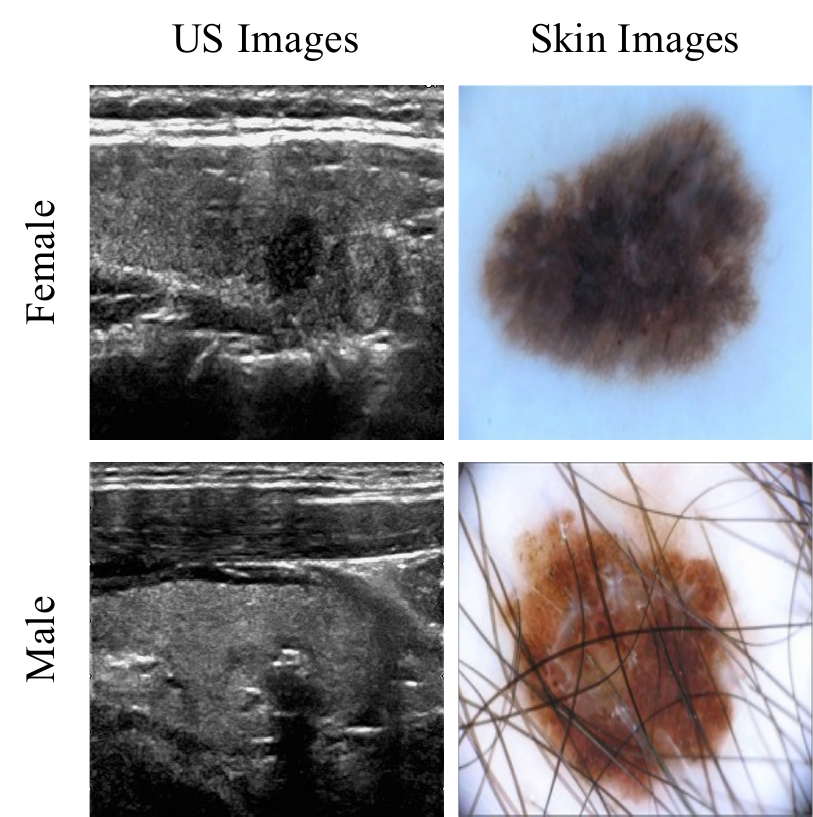}
    \caption{Male and Female images in an US dataset and a skin lesion dataset. Compared to skin, US images are more difficult to distinguish the sex of the image.}
    \label{fig:visual}
\end{figure}

\subsection{Quntitative Evaluation}
\textbf{TUSC Dataset:}
Among the seven methods evaluated on the TUSC, the average Dice score is approximately 68\%. CMU-Net achieves the highest performance with a Dice score of 74.41\%, while U-Net demonstrates the least favorable performance. In terms of fairness criteria, CMU-Net exhibits the best ESSP for both \textit{Sex} and \textit{Age}, which can be attributed to its high segmentation utility. Conversely, UNeXt records the lowest DD and EOpp for \textit{Sex}, and the lowest EOpp for \textit{Age}.

\noindent{\textbf{TUS Dataset:}}
The trends are more distinct for the seven methods on the TUS. CMU-Net and Attention U-Net demonstrate the best and second-best performance, as well as fairness, across almost all metrics, followed by TransUnet and U-Net++. Conversely, U-Net, UNeXt, and CMUNeXt exhibit higher levels of unfairness on both \textit{Sex} and \textit{Age}.

\subsection{Statistical Evaluation}



From Fig.~\ref{fig.1}, it is evident that, across all seven algorithms evaluated on the two datasets, the performance of the female group surpasses that of the male group. Additionally, the performance of the young group is consistently higher than that of the male group. The uniform performance disparities concerning these two sensitive attributes underscore varying degrees of unfairness across all seven algorithms.

To delve deeper, a Mann-Whitney U test is conducted on the Dice scores of the seven algorithms for the two sensitive attributes on the test set. Notably, U-Net, the foundational model for other sex-specific methods, exhibits statistically significant performance disparities on \textit{Sex} in the TUSC dataset and on \textit{Age} in both TUSC and TUS datasets. Furthermore, TransUnet (on TUSC), Attention U-Net, and CMUNet (on TUS) show distinguishable performances on the Female and Male groups ($0.01<p\leq0.05$). UNeXt demonstrates significant unfairness on \textit{Age} (on TUSC). All algorithms, except for UNeXt and UNet++, exhibit significant unfairness on the \textit{Age} attribute in the TUS dataset.

In conclusion, the presence of model unfairness is evident across all seven models, as observed through both quantitative evaluation and statistical analysis. This sheds light on the importance of considering fairness alongside the segmentation utility of deep learning models.


\section{Discussion}

To the best of our knowledge, this work represents the first comprehensive analysis and comparison of fairness issues in US segmentation tasks. Through the computation of three fairness metrics and rigorous statistical hypothesis testing, we reveal that all seven state-of-the-art U-Net-based segmentation models exhibit varying degrees of unfairness on the two US datasets.

This outcome is unexpected, considering that prior research on fairness typically focused on tasks where sensitive attributes could be easily distinguished. For instance, in skin lesion classification, differences in skintone are readily apparent. Contrary to these tasks, US images present a more challenging scenario for distinguishing between male and female, as illustrated in Fig.~\ref{fig:visual}.

Given that most fairness research in medical image analysis has been conducted on classification tasks, where acknowledged fairness metrics exist, the evaluation of unfairness in segmentation tasks, especially US segmentation, requires careful consideration. While some studies adopt definitions from classification tasks, applying metrics like the disparity of Dice or Hausdorff Distance in segmentation, the unique challenges of evaluating unfairness in US segmentation demand further exploration.

Our study has certain limitations. We independently consider Sex and Age, without evaluating the intersection between these two variables, which may introduce higher levels of unfairness due to the complex relationships among segmentation targets, sex, and age. Additionally, assessing model unfairness based on the actual age of patients, rather than a binary categorical label, is an essential step in the analysis process.

In conclusion, the experiments in this paper uncover the presence of unfairness in US segmentation tasks. Therefore, careful use of these deep learning models is imperative to avoid producing unfair diagnoses. Future work should focus on designing fairness mitigation algorithms tailored to address this specific challenge in US segmentation.

\section{Conclusion}

In this paper, we present the inaugural in-depth evaluation of fairness issues in US segmentation tasks, assessing seven state-of-the-art DL-based segmentation models. The results reveal varying degrees of unfairness across all these methods, underscoring the importance of considering model unfairness alongside utility. This not only contributes to enhancing the reliability and interpretability of deep learning-based models but also serves as a critical step in preventing the unequal treatment of underprivileged patients.

\clearpage
\bibliographystyle{IEEEbib}
\bibliography{strings,refs}

\begin{thebibliography}{10}

\bibitem{zhou2021review}
S~Kevin Zhou, Hayit Greenspan, Christos Davatzikos, James~S Duncan, Bram Van~Ginneken, Anant Madabhushi, Jerry~L Prince, Daniel Rueckert, and Ronald~M Summers,
\newblock ``A review of deep learning in medical imaging: Imaging traits, technology trends, case studies with progress highlights, and future promises,''
\newblock {\em Proceedings of the IEEE}, vol. 109, no. 5, pp. 820--838, 2021.

\bibitem{baltruschat2019comparison}
Ivo~M Baltruschat, Hannes Nickisch, Michael Grass, Tobias Knopp, and Axel Saalbach,
\newblock ``Comparison of deep learning approaches for multi-label chest x-ray classification,''
\newblock {\em Scientific reports}, vol. 9, no. 1, pp. 6381, 2019.

\bibitem{akkus2017deep}
Zeynettin Akkus, Alfiia Galimzianova, Assaf Hoogi, Daniel~L Rubin, and Bradley~J Erickson,
\newblock ``Deep learning for brain mri segmentation: state of the art and future directions,''
\newblock {\em Journal of digital imaging}, vol. 30, pp. 449--459, 2017.

\bibitem{larrazabal2020gender}
Agostina~J Larrazabal, Nicol{\'a}s Nieto, Victoria Peterson, Diego~H Milone, and Enzo Ferrante,
\newblock ``Gender imbalance in medical imaging datasets produces biased classifiers for computer-aided diagnosis,''
\newblock {\em Proceedings of the National Academy of Sciences}, vol. 117, no. 23, pp. 12592--12594, 2020.

\bibitem{cmunet}
Fenghe Tang, Lingtao Wang, Chunping Ning, Min Xian, and Jianrui Ding,
\newblock ``Cmu-net: A strong convmixer-based medical ultrasound image segmentation network,''
\newblock {\em arXiv preprint arXiv:2210.13012}, 2022.

\bibitem{glocker2022risk}
Ben Glocker, Charles Jones, Melanie Bernhardt, and Stefan Winzeck,
\newblock ``Risk of bias in chest x-ray foundation models,''
\newblock {\em arXiv preprint arXiv:2209.02965}, 2022.

\bibitem{deng2023fairness}
Wenlong Deng, Yuan Zhong, Qi~Dou, and Xiaoxiao Li,
\newblock ``On fairness of medical image classification with multiple sensitive attributes via learning orthogonal representations,''
\newblock {\em arXiv preprint arXiv:2301.01481}, 2023.

\bibitem{xu2023fairadabn}
Zikang Xu, Shang Zhao, Quan Quan, Qingsong Yao, and S~Kevin Zhou,
\newblock ``Fairadabn: Mitigating unfairness with adaptive batch normalization and its application to dermatological disease classification,''
\newblock {\em arXiv preprint arXiv:2303.08325}, 2023.

\bibitem{lee2023investigation}
Tiarna Lee, Esther Puyol-Ant{\'o}n, Bram Ruijsink, Keana Aitcheson, Miaojing Shi, and Andrew~P King,
\newblock ``An investigation into the impact of deep learning model choice on sex and race bias in cardiac mr segmentation,''
\newblock in {\em Workshop on Clinical Image-Based Procedures}. Springer, 2023, pp. 215--224.

\bibitem{zong2022medfair}
Yongshuo Zong, Yongxin Yang, and Timothy Hospedales,
\newblock ``Medfair: Benchmarking fairness for medical imaging,''
\newblock {\em arXiv preprint arXiv:2210.01725}, 2022.

\bibitem{unet}
Olaf Ronneberger, Philipp Fischer, and Thomas Brox,
\newblock ``U-net: Convolutional networks for biomedical image segmentation,''
\newblock in {\em Medical Image Computing and Computer-Assisted Intervention--MICCAI 2015: 18th International Conference, Munich, Germany, October 5-9, 2015, Proceedings, Part III 18}. Springer, 2015, pp. 234--241.

\bibitem{unext}
Jeya Maria~Jose Valanarasu and Vishal~M Patel,
\newblock ``Unext: Mlp-based rapid medical image segmentation network,''
\newblock in {\em International Conference on Medical Image Computing and Computer-Assisted Intervention}. Springer, 2022, pp. 23--33.

\bibitem{unet++}
Zongwei Zhou, Md~Mahfuzur~Rahman Siddiquee, Nima Tajbakhsh, and Jianming Liang,
\newblock ``Unet++: Redesigning skip connections to exploit multiscale features in image segmentation,''
\newblock {\em IEEE transactions on medical imaging}, vol. 39, no. 6, pp. 1856--1867, 2019.

\bibitem{attunet}
Ozan Oktay, Jo~Schlemper, Loic~Le Folgoc, Matthew Lee, Mattias Heinrich, Kazunari Misawa, Kensaku Mori, Steven McDonagh, Nils~Y Hammerla, Bernhard Kainz, et~al.,
\newblock ``Attention u-net: Learning where to look for the pancreas,''
\newblock {\em arXiv preprint arXiv:1804.03999}, 2018.

\bibitem{cmunext}
Fenghe Tang, Jianrui Ding, Lingtao Wang, Chunping Ning, and S~Kevin Zhou,
\newblock ``Cmunext: An efficient medical image segmentation network based on large kernel and skip fusion,''
\newblock {\em arXiv preprint arXiv:2308.01239}, 2023.

\bibitem{transunet}
Jieneng Chen, Yongyi Lu, Qihang Yu, Xiangde Luo, Ehsan Adeli, Yan Wang, Le~Lu, Alan~L Yuille, and Yuyin Zhou,
\newblock ``Transunet: Transformers make strong encoders for medical image segmentation,''
\newblock {\em arXiv preprint arXiv:2102.04306}, 2021.

\bibitem{TUSC}
Stanford~AIMI shared datasets,
\newblock ``Thyroid ultrasound cine-clip,'' \url{https://stanfordaimi.azurewebsites.net/datasets/a72f2b02-7b53-4c5d-963c-d7253220bfd5}.

\bibitem{tian2023fairseg}
Yu~Tian, Min Shi, Yan Luo, Ava Kouhana, Tobias Elze, and Mengyu Wang,
\newblock ``Fairseg: A large-scale medical image segmentation dataset for fairness learning with fair error-bound scaling,''
\newblock {\em arXiv preprint arXiv:2311.02189}, 2023.

\bibitem{hardt2016equality}
Moritz Hardt, Eric Price, and Nati Srebro,
\newblock ``Equality of opportunity in supervised learning,''
\newblock {\em Advances in neural information processing systems}, vol. 29, 2016.

\end{thebibliography}

\end{document}